\documentclass[10pt,onecolumn,letterpaper]{article}

\usepackage{times}
\usepackage{epsfig}
\usepackage{graphicx}
\usepackage{amsmath}
\usepackage{amssymb}
\usepackage{array}
\usepackage{adjustbox}
\usepackage{array}
\usepackage{cellspace}
\usepackage{hhline}

\graphicspath{{images/}}
\usepackage[utf8]{inputenc} 
\usepackage{multirow}
\usepackage{placeins}
\usepackage{lineno}
\usepackage{color}
\usepackage{hyperref}
\usepackage{subcaption}
\pdfoutput=1
\newcommand{\DBGOTPD}{\texttt{GOTPD1+}}
\newcommand{\DBZHANG}{\texttt{Zhang2012}}
\newcommand{\DBMIVIA}{\texttt{MIVIA}}

\newcommand{\WaterFilling}{\texttt{WaterFilling}}
\newcommand{\SLICEPCA}{\texttt{SLICE$_{PCA}$}}
\newcommand{\SLICESVM}{\texttt{SLICE$_{SVM}$}}
\newcommand{\MexicanHat}{\texttt{MexicanHat}}
\newcommand{\DPDNet}{\texttt{DPDnet}}
\newcommand{\DPDNetFast}{\texttt{DPDnet$_{fast}$}}

\newcommand{\mbf}{\mathbf}

\definecolor{greenao}{rgb}{0.0, 0.5, 0.0}

\newcommand*{\mathcolor}{}
\def\mathcolor#1#{\mathcoloraux{#1}}
\newcommand*{\mathcoloraux}[3]{  \protect\leavevmode
	\begingroup
	\color#1{#2}#3  \endgroup
}

\definecolor{111}{rgb}{0,1,0}
\definecolor{00}{rgb}{1.00000000000000000000,.49803921568627450980,0}
\definecolor{222}{rgb}{0.8,0.9568627451,0.56470588235}
\definecolor{111}{rgb}{0.8,0.9568627451,0.56470588235}

\usepackage[table]{xcolor}

\begin{document}
\title{DPDnet: A Robust People Detector using Deep
	Learning with an Overhead Depth Camera}

\author{David Fuentes-Jimenez,  Roberto Martin-Lopez,\\
	 Cristina Losada-Gutierrez, David Casillas-Perez,\\
	  Javier Macias-Guarasa, Daniel Pizarro, Carlos A.Luna\\
Universidad de Alcal\'a\\
{\tt\small \{d.fuentes, david.casillas, roberto.martin\}@edu.uah.es}\\
{\tt\small\{daniel.pizarro, cristina.losada,javier.maciasguarasa\}@uah.es} }
\maketitle

\begin{abstract}
In this paper we propose a method based on deep learning that detects multiple people from a single overhead depth image with high reliability. Our neural network, called \DPDNet{}, is based on two fully-convolutional \emph{encoder-decoder} neural blocks based on
residual layers. The \emph{main block} takes a depth image as input and generates a pixel-wise confidence map, where each detected person in the image is represented by a Gaussian-like distribution. The \emph{refinement block} combines the depth image and the output from the main block, to refine the confidence map. Both blocks are simultaneously trained end-to-end using depth images and head position labels. 
	
The experimental work shows that \DPDNet{} outperforms
state-of-the-art methods, with accuracies greater than 99\% in three
different publicly available datasets, without retraining not
fine-tuning. In addition, the computational complexity of our
proposal is independent of the number of people in the scene and
runs in real time using conventional GPUs.
\end{abstract}

\section{Introduction and State of the Art}
\label{sec:introduction}

People detection and tracking have attracted a growing interest of the
scientific community in recent years because of its applications in
multiple areas such as video-surveillance, access control or human
behavior analysis. Most of these applications require robust and
non-invasive systems (\emph{i.e} without adding turnstiles or other
contact systems).  Consequently, there have been an increasing number of
works in the literature that address people detection and
tracking~\cite{nguyen2016,brunetti2018,zhang2018,stewart2016,gao2016,du2017,yunfei2018}
using cameras and other non-invasive sensors. Despite the large amount
of existing works addressing this task, it still presents open
challenges~\cite{zhang2016far} in terms of accuracy, stability and
computational complexity, and specially in highly populated scenes.

The first proposed works in non-invasive people detection were based on
the use of RGB cameras. To name a few,~\cite{Ramanan2007} proposed a
system based on learning person appearance models,
whereas~\cite{Andriluka2008} used hierarchical Gaussian Process Latent
Variable Models (hGPLVM) for modeling people. Other approaches were
based on face detection~\cite{TsongYi2010} or interest point
classification~\cite{ChiYoon2013}.  In recent years, improvements in
technology and the availability of large annotated image datasets such
as Imagenet~\cite{imagenet2012} have allowed Deep Neural Networks
(DNNs) to become the state-of-the-art solutions for tasks such as object
detection~\cite{yolo2016}, segmentation~\cite{cao2017} and
classification~\cite{hoo2016} in RGB images. DNNs have also been
proposed for the specific task of people detection. In particular,
~\cite{ouyang2013} proposes a new DNN architecture that jointly performs
feature extraction, deformation handling, occlusion handling and
classification. In~\cite{wang2017} they propose a Convolutional Neural
Network (CNN) as a feature extractor followed by a proposal selection
network.  In~\cite{du2017}, they combine a novel DNN model that jointly
optimizes pedestrian detection with semantic tasks, including both
pedestrian and scene attributes. Finally,~\cite{zhao2017} uses a novel
physical radius-depth (PRD) to detect human candidates, followed by a
CNN for human feature extraction. These proposals present good results
in controlled conditions, but all of them have problems in scenarios
with a high amount of clutter.

To reduce the effect of occlusions and to improve detection accuracy,ºa
some works propose the fusion of RGB and depth
data~\cite{liu2015,murano2015,zhou2017,ren2017,cao2017}. Other works
propose to place cameras in overhead
configurations~\cite{Antic2009,Zebin2014,ByoungKyu2012,delpizzo2016counting}.

A key factor in some applications is to develop detection methods that
also preserve privacy, preventing anyone from using the system to find
out the identity of each person in the scene.  Taking this factor into
account, in the last few years several works have appeared that detect
people by using sensors and camera configurations that do not easily
allow to identify any person being detected. For
example,~\cite{Chan2008} propose the use of a low resolution camera that
is located far away from the users. This setup limits the applicability
of the proposed method to particular scenarios and makes the system weak
against occlusions.  Other works employ depth sensors for people
detection, either based on Time-of-Flight
(ToF)~\cite{Bevilacqua2006,Stahlschmidt2013,Stahlschmidt2014,Li2014} or
structured
light~\cite{Zhang2012,galvcik2013real,rauter2013reliable,zhu2013human,delpizzo2016counting,vera2016counting}
technologies.  All these works use an overhead camera to reduce the
occlusions.

Some of these methods are based on finding distinctive maxima of the
depth image~\cite{Bevilacqua2006,Zhang2012}, or a filtered depth image
using the normalized Mexican Hat Wavelet
filter~\cite{Stahlschmidt2013,Stahlschmidt2014}. However, they usually
fail to separate clusters of people that are close to each other in the
scene or produce false detections when parts of the body, other than the
head, are closer to the camera.  Moreover, the proposals
by~\cite{Zhang2012}, and~\cite{Stahlschmidt2013,Stahlschmidt2014} do not
include a classification stage, so that they are not able to
discriminate people from other objects in the scene.

In order to reduce false positives, other works include a classification
stage that allows discriminating between people and other elements in
the scene. In
particular,~\cite{galvcik2013real,vera2016counting,zhu2013human,luna2017}
use shape descriptors that encode the morphology of the human head and
shoulders as seen in the depth image. These proposals are able to
discriminate between people and other elements in the scene, but in some
of them, their detection rates significantly decrease if people are
close to each other.

\begin{figure}[h]
	\begin{center}
		\begin{subfigure}[t]{0.49\textwidth}
			\includegraphics[width=\textwidth]{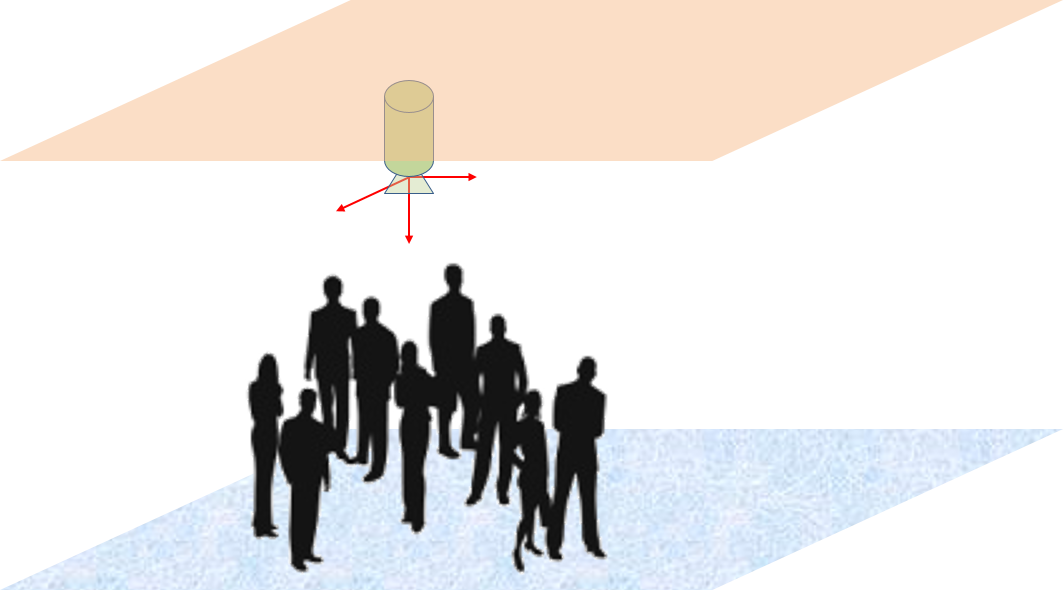}
			\caption{Camera location above sensed area.}
			\label{fig:tof-scene-general}
		\end{subfigure}
		\;    \begin{subfigure}[t]{0.49\textwidth}
			\includegraphics[width=\textwidth]{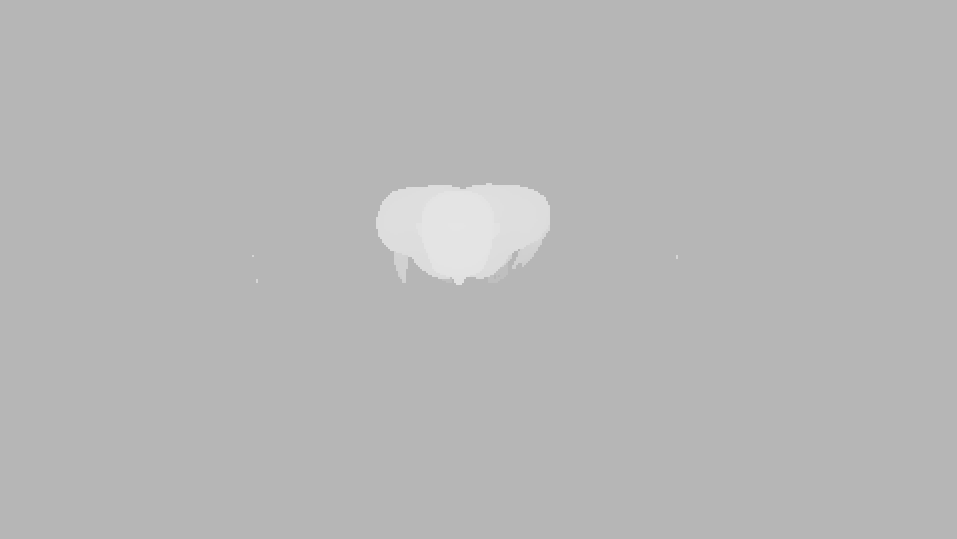}
			\caption{Example of recorded frame in sensed area.}
			\label{fig:frame_camara2}
		\end{subfigure}
	\end{center}
	\caption{Scheme of camera location above sensed area and example of
		recorded frame}
	\label{fig:tof-scene-general-and-sample-frame}
\end{figure}

In this paper, we propose a method based on DNNs, called \DPDNet{}, for
robust and reliable detection of multiple people from depth images
acquired using an overhead depth camera, as shown in
Figure~\ref{fig:tof-scene-general-and-sample-frame}. The proposed neural
network is trained end-to-end using depth images and the annotated
position for each person in the scene.  Once trained, \DPDNet{} is able
to discriminate between people and other elements in the scene, even if
the test images are acquired using a different sensor from the one used
during training, or if it is located at a different height.

In addition to the evaluation done on a specifically recorded and labeled
database~\cite{gotpd1-2016}, our method has been tested using two
widely used depth image datasets~\cite{Zhang2012,mivia2016}
without fine tuning or retraining, significantly improving the results
of previous works in the literature. \DPDNet{} runs in real time using
conventional GPUs, and its computational requirements do not depend on
the number of people in the scene.  To the best of the authors
knowledge, there are no other works in the literature able to detect
people in depth images with comparable accuracy and reliability.

The structure of the paper is described as follows:
section~\ref{sec:tofnet} describes the person detection algorithm,
section~\ref{sec:experimental-work} includes the experimental setup,
results and discussion, and section~\ref{sec:conclusions} contains the
main conclusions and some ideas for future work.

\section{DPDnet People Detector}
\label{sec:tofnet}

\subsection{Problem Formulation}
\label{Problem Formulation}

\begin{figure}[h]
	\centering
	\includegraphics[width=1.0\linewidth]{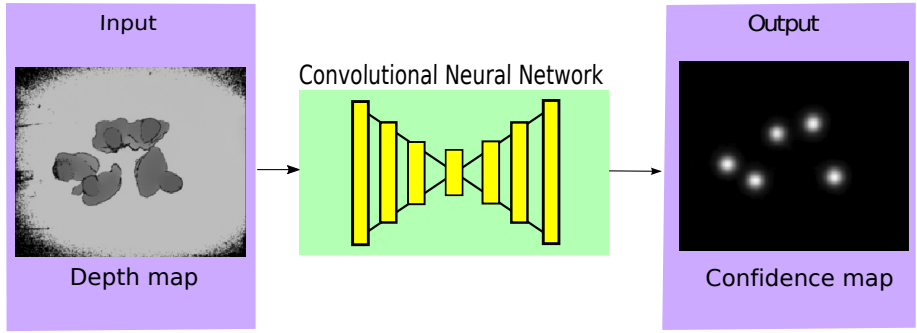}
	\caption{Depth input image and output confidence map.}
	\label{fig:inputOutput1}.
\end{figure}

We propose the \DPDNet{} neural network to detect multiple people from a
single depth image. The output consists of a multi-modal confidence map,
the same size as the input image, that jointly codifies the detection of
multiple people in scene (see Figure \ref{fig:inputOutput1}).

The confidence map assigns a 2D Gaussian distribution centered in each
detected person. We use the centroid of the head in image coordinates as
the localization reference. The 2D position of each person can be easily
obtained from the confidence map by detecting its local maxima.

The advantages of this strategy are twofold: first, learning the
confidence map is a well defined process to detect multiple hypothesis
as opposed to using other parametrizations (\textit{e.g} multiple 2D
coordinate vectors) which can lead to ambiguities in the regression
function. Second, it makes computational complexity to be independent
from the number of people in the scene, assuming that the time used for
extracting multiple detection hypotheses from the confidence map is
mostly negligible.

\subsection{DPDnet Network Architecture}
\label{sec:Arquitecture}

\begin{figure*}[h]
	\centering
	\includegraphics[width=1.0\linewidth]{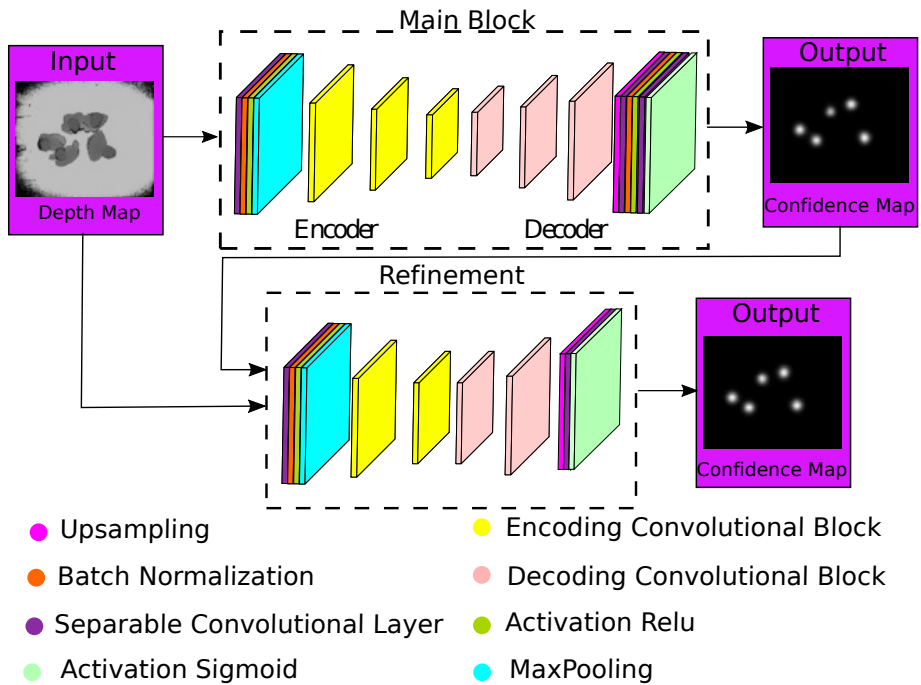}
	\caption{The \DPDNet{} architecture divided into the \emph{main block}
		and the \emph{refinement block}. Figure shows the general
		\emph{encoder-decoder} structure in both blocks. }
	\label{fig:diagramaTOF}.
\end{figure*}

Figure~\ref{fig:diagramaTOF} illustrates our approach to detect people
from a single depth image.

Our \DPDNet{} neural network consists of two blocks: \emph{the main
	block} and the \emph{refinement block}. Both blocks use the
\emph{encoder-decoder} architecture inspired in the
\emph{Segnet}~\cite{Segnet} model, originally proposed in semantic
segmentation, and use the residual layers proposed in the
\emph{ResNet}~\cite{resnet16} model. 

The \emph{main block} takes a single depth image as input, and outputs a
confident map. The \emph{refinement block} takes as inputs the input
depth image along with the confidence map obtained from the \emph{main
	block}, and outputs a refined confidence map. In
Section~\ref{sec:experimental-work}, we show that the \emph{refinement
	block} significantly outperforms the results obtained by the
\emph{main block} alone. Similar refinement blocks have been proposed
for other tasks, such as in monocular reconstruction
methods~\cite{monocular_reconstruction}.

\begin{figure}[h]
	\centering
	\includegraphics[width=1\linewidth]{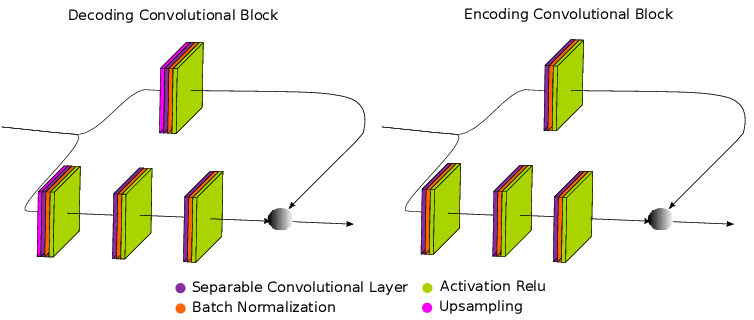}
	\caption{DPDnet decoding and encoding block. }
	\label{fig:encodingDecodingBlock}
\end{figure}

The structure of the \emph{main block} is detailed in
Table~\ref{tb:mainBlock}. It takes $212\times256$ depth images as input,
and uses separable convolutional layers based on the
Xception~\cite{Xception} model. Separable convolutions are faster than
conventional convolutions, while achieving similar performance. In
addition to this, we use ReLU activations~\cite{agarap2018deep} and
batch normalization. To build the encoder-decoder core of the \emph{main
	block}, we use several units of residual blocks, similar to those
proposed in the ResNet~\cite{resnet16} model. Encoding blocks downsample
the spatial dimensions by a factor of two, and decoding blocks use
upscaling layers to increase the spatial dimensions by the same
factor. Their basic architecture is described in
Figure~\ref{fig:encodingDecodingBlock} and Tables \ref{tb:Eblock} and
\ref{tb:Dblock}.

\begin{table}
	\centering
	\caption{Encoding Block  }
	\begin{tabular}{ l | c | c}
		\hline
		\multicolumn{3}{c}{\textbf{Encoding Block (filters=(a,b,c)),kernel size=k}} \\
		\hline
		\hline
		\textbf{Layer} & \textbf{Parameters} & \textbf{Output Size} \\
		\hline
		\hline
		Input & - & (Width, Height, Depth)\\
		\hline
		Convolution (Main Branch) & \begin{tabular}{@{}c@{}}kernel=(1, 1) \\ strides=(2, 2) \\ filters=a  \end{tabular}& (Width/2, Height/2, a)\\
		\hline
		Batch Normalization (Main Branch) & \multicolumn{2}{c}{-} \\
		\hline
		Activation (Main Branch)& \multicolumn{2}{c}{ReLU} \\
		\hline
		Convolution (Main Branch) & \begin{tabular}{@{}c@{}}kernel=(k, k) \\ strides=(1, 1) \\ filters=b  \end{tabular} & (Width/2, Height/2, b)\\
		\hline
		Batch Normalization (Main Branch)& \multicolumn{2}{c}{-} \\
		\hline
		Activation (Main Branch)& \multicolumn{2}{c}{ReLU} \\
		\hline
		Convolution (Main Branch) & \begin{tabular}{@{}c@{}}kernel=(1, 1) \\ strides=(1, 1) \\ filters=c  \end{tabular} & (Width/2, Height/2, c)\\
		\hline
		Batch Normalization (Main Branch)& \multicolumn{2}{c}{-} \\
		\hline	
		Convolution (Shortcut) &\begin{tabular}{@{}c@{}}kernel=(1, 1) \\ strides=(2, 2) \\ filters=c  \end{tabular}& (Width/2, Height/2, c)\\
		\hline
		Batch Normalization (Shortcut)& \multicolumn{2}{c}{-} \\
		\hline
		Add (Main Branch+Shortcut)& - & (Width/2, Height/2, a)\\
		\hline
		Activation (Main Branch+Shortcut)&\multicolumn{2}{c}{ReLU} \\
		\hline
	\end{tabular}
	\label{tb:Eblock}
\end{table}
\begin{table}
	\centering
	\caption{Decoding Block}
	\begin{tabular}{ l | c | c}
		\hline
		\multicolumn{3}{c}{\textbf{Decoding Block (filters=(a,b,c)),kernel size=k}} \\
		\hline
		\hline
		\textbf{Layer} & \textbf{Parameters} & \textbf{Output Size} \\
		\hline
		\hline
		Input & - & (Width, Height, Depth)\\
		\hline
		Upsampling (Main Branch) & size=(2,2) & (2*Width,2*Height,Depth) \\
		\hline
		Convolution (Main Branch) &  \begin{tabular}{@{}c@{}}kernel=(1, 1) \\ strides=(1, 1) \\ filters=a  \end{tabular} & (Width/2, Height/2, a)\\
		\hline
		Batch Normalization (Main Branch) & \multicolumn{2}{c}{-} \\
		\hline
		Activation (Main Branch)& \multicolumn{2}{c}{ReLU} \\
		\hline
		Convolution (Main Branch) & \begin{tabular}{@{}c@{}}kernel=(k, k) \\ strides=(1, 1) \\ filters=b  \end{tabular}& (Width/2, Height/2, b)\\
		\hline
		Batch Normalization (Main Branch)& \multicolumn{2}{c}{-} \\
		\hline
		Activation (Main Branch)& \multicolumn{2}{c}{ReLU} \\
		\hline
		Convolution (Main Branch) &\begin{tabular}{@{}c@{}}kernel=(1, 1) \\ strides=(1, 1) \\ filters=c  \end{tabular} & (Width/2, Height/2, c)\\
		\hline
		Batch Normalization (Main Branch)& \multicolumn{2}{c}{-} \\
		\hline
		Upsampling (Shortcut) & size=(2,2) & (2*Width,2*Height,Depth) \\
		\hline	
		Convolution (Shortcut) & \begin{tabular}{@{}c@{}}kernel=(1, 1) \\ strides=(1, 1) \\ filters=c  \end{tabular}& (Width/2, Height/2, c)\\
		\hline
		Batch Normalization (Shortcut)& \multicolumn{2}{c}{-} \\
		\hline
		Add (Main Branch+Shortcut)& - & (Width/2, Height/2, a)\\
		\hline
		Activation (Main Branch+Shortcut)&\multicolumn{2}{c}{ReLU} \\
		\hline
	\end{tabular}
	\label{tb:Dblock}
\end{table}

Finally, the last block of the CNN and the cropping layers adapt the
output of the last decoding block to the output size of the confidence
map ($212\times256$), which is the same size as the input
image. Table~\ref{tb:mainBlock} summarizes all layers that appear in the
\emph{main block}.

\begin{table}
	\centering
	\caption{Main Stage. In Encoding and Decoding Blocks, a, b and c represents the number of filters of the intermediate convolutional layers in the branches.}
	\begin{tabular}{ l | c | c  }
		\hline
		\multicolumn{3}{c}{\textbf{Main Block}} \\
		\hline
		\hline
		\multicolumn{1}{c}{\textbf{Layer}} & \textbf{Output Size} & \textbf{Parameters} \\
		\hline
		\hline
		Input & 212x256x1 & - \\
		\hline
		Convolution & 106x128x64 & kernel=(7, 7) / strides=(2, 2) \\
		\hline
		Batch Normalization & \multicolumn{2}{c}{-} \\
		\hline
		Activation & \multicolumn{2}{c}{ReLU} \\
		\hline
		Max Pooling & 35x42x64 & size=(3, 3) \\
		\hline
		\hline
		Encoding Conv. Block & 35x42x256 & \begin{tabular}{@{}c@{}}kernel=(3, 3) / strides=(1, 1) \\ (a=64, b=64, c=256) \end{tabular} \\
		\hline
		Encoding Conv. Block & 18x21x512 & \begin{tabular}{@{}c@{}}kernel=(3, 3) / strides=(2, 2) \\ (a=128, b=128, c=512)  \end{tabular} \\
		\hline
		Encoding Conv. Block & 9x11x1024 & \begin{tabular}{@{}c@{}}kernel=(3, 3) / strides=(2, 2) \\ (a=256, b=256, c=1024)  \end{tabular} \\
		\hline
		\hline
		Decoding Separable Conv. Block & 9x11x256 & \begin{tabular}{@{}c@{}}kernel=(3, 3) / strides=(1, 1) \\ (a=1024, b=1024, c=256)  \end{tabular} \\
		\hline
		Decoding Separable Conv. Block & 18x22x128 & \begin{tabular}{@{}c@{}}kernel=(3, 3) / strides=(2, 2) \\ (a=512, b=512, c=128)  \end{tabular} \\
		\hline
		Decoding Separable Conv. Block & 36x44x64 & \begin{tabular}{@{}c@{}}kernel=(3, 3) / strides=(2, 2) \\ (a=256, b=256, c=64)  \end{tabular} \\
		\hline
		\hline
		Cropping & 36x43x64 & cropping=[(2, 2) (1, 1)] \\
		\hline
		Up Sampling & 108x129x64 & size=(3, 3) \\
		\hline
		Convolution & 216x258x64 & kernel=(7, 7) / strides=(2, 2) \\
		\hline
		Cropping & 212x256x64 & cropping=[(2, 2) (1, 1)] \\
		\hline
		Batch Normalization & \multicolumn{2}{c}{-} \\
		\hline
		Activation & \multicolumn{2}{c}{ReLU} \\
		\hline
		Convolution & 212x256x1 & kernel=(3, 3) / strides=(1, 1) \\
		\hline
		Activation & \multicolumn{2}{c}{Sigmoid} \\
		\hline
		Output 1 & 212x256x1 & - \\
		\hline	
	\end{tabular}
	\label{tb:mainBlock}
\end{table}

The \emph{refinement stage} is a reduced version of the \emph{main
	block} with two encoding and decoding blocks. It is thus more shallow
than the \emph{main block}. The \emph{refinement block} takes as inputs
the input depth image and the confidence map computed from the
\emph{main block}, and outputs a new ``refined'' confidence
map. 
Table~\ref{tb:refinement} summarizes all the
layers and blocks used in the \emph{refinement stage}.

\begin{table}
	\centering
	\caption{Refinement Stage }
	\begin{tabular}{ l | c | c  }
		\hline
		\multicolumn{3}{c}{\textbf{Refinement Block}} \\
		\hline
		\hline
		\multicolumn{1}{c}{\textbf{Layer}} & \textbf{Output Size} & \textbf{Parameters} \\
		\hline
		\hline
		Output 1 + Input & 212x256x1 & - \\
		\hline
		Convolution & 106x128x64 & kernel=(7, 7) / strides=(2, 2) \\
		\hline
		Batch Normalization & \multicolumn{2}{c}{-} \\
		\hline
		Activation & \multicolumn{2}{c}{ReLU} \\
		\hline
		Max Pooling & 35x42x64 & size=(3, 3) \\
		\hline
		\hline
		Encoding Conv. Block & 35x42x256 & \begin{tabular}{@{}c@{}}kernel=(3, 3) / strides=(1, 1) \\ (a=64, b=64, c=256) \end{tabular} \\
		\hline
		Encoding Conv. Block & 18x21x512 & \begin{tabular}{@{}c@{}}kernel=(3, 3) / strides=(2, 2) \\ (a=128, b=128, c=512)  \end{tabular} \\
		\hline
		\hline
		Decoding Conv. Block & 36x42x128 & \begin{tabular}{@{}c@{}}kernel=(3, 3) / strides=(2, 2) \\ (a=512, b=512, c=128)  \end{tabular} \\
		\hline
		Decoding Conv. Block & 72x84x64 & \begin{tabular}{@{}c@{}}kernel=(3, 3) / strides=(2, 2) \\ (a=256, b=256, c=64)  \end{tabular} \\
		\hline
		\hline
		Zero Padding & 72x86x64 & padding=(0, 1) \\
		\hline
		Up Sampling & 216x258x64 & size=(3, 3) \\
		\hline
		Cropping & 212x256x64 & cropping=[(2, 2) (1, 1)] \\
		\hline
		Convolution & 212x256x1 & kernel=(3, 3) / strides=(1, 1) \\
		\hline
		Activation & \multicolumn{2}{c}{Sigmoid} \\
		\hline
		Output 2 & 212x256x1 & - \\
		\hline	
	\end{tabular}
	\label{tb:refinement}
\end{table}

In addition to the \DPDNet{} architecture described above, we have
adapted the same model to work with images half the resolution of the
original ones ($106\times 128$), seamlessly producing downscaled
confident maps. We will refer to this reduced model as \DPDNetFast{} in
section~\ref{sec:experimental-work}.  The reduction of input and output
size allows low cost \emph{GPUs} to run our model with a very small
impact in terms of accuracy.

\subsection{Training procedure}
\label{sec:Training}

We denote by $p_i$ the input depth image and by $q_i$ the target
confident map, where $p_i,q_i \in\mathbb{R}^{212\times 256}$. We assume
that both $p_i$ and $q_i$ are normalized in the interval $[0,1]$. The
target confidence map $q_i$ is generated by a sum of two-dimensional
Gaussian distributions with covariance matrix
$P=\sigma^2 \mathbb{I}_{2\times 2}$ and centered around the labeled
position of each person in the scene. We use $\sigma=6$ pixels in all
our experiments.

The \emph{main} and \emph{refinement} blocks of the \DPDNet{} are
defined with the functions $\mathcal{T}_m(p_i,\theta_m)$ and
$\mathcal{T}_r(p_i,q_i,\theta_r)$ respectively. All trainable parameters
of the \emph{main block} are defined with vector $\theta_m$, and
similarly with $\theta_r$ for the \emph{refinement block}. When
referring to all trainable parameters of the entire \emph{DPDnet} model
we use their concatenation $\theta=(\theta_m,\theta_r)$.

The complete \emph{DPDnet} function $\mathcal{T}(p_i,\theta)$ is
obtained by composition of $\mathcal{T}_m$ and $\mathcal{T}_r$:
\begin{equation}
\mathcal{T}(p_i,\theta) = \mathcal{T}_r\left( p_i,\mathcal{T}_m(p_i,\theta_m),\theta_r \right)
\end{equation}

We train our network using a set of corresponding depth images and
confidence maps $\{(p_i,q_i)\}_{i=1,\dots,N}$. The following loss
function is used:

\begin{equation}
\mathcal{L}(\theta) = \frac{1}{N}\sum_{i=1}^N \|\mathcal{T}(p_i,\theta) - q_{i}\|^2 + \lambda \frac{1}{N}\sum_{i=1}^N \|\mathcal{T}_m(p_i,\theta_m) - q_{i}\|^2,
\label{eq:1}
\end{equation}
where $\lambda$ is a hyper-parameter that weights the importance of the
two terms in the loss function and is set to $\lambda=1$ in our
experiments. Note that the second term of equation (\ref{eq:1}) imposes
the \emph{main block} to be as close as possible to the target
confidence map. This forces the \emph{refinement block} to improve the
output of the \emph{main block}.

By minimizing equation (\ref{eq:1}), we train the entire \DPDNet{} model
end-to-end in a single step without separating the \emph{main block} and
\emph{refinement block}.

In our experiments, we train the network for $50$ epochs and use a
validation set to select the best model. We employ the \emph{Adam}
optimizer~\cite{kingma2014adam} with $0.001$ as initial learning
rate. This optimizer has been chosen because of its adaptive
capabilities in terms of learning rate. The great advantage of Adam is
the fact that it starts from an initial learning rate set by the user
and then uses the first and second moments of the gradient to adapt the
learning rate to the situation stipulated by Adam in the loss
function. This provides unrivaled speed and robustness compared to
other optimizers, making Adam a good choice for the problem proposed
here.

\section{Experimental Work}
\label{sec:experimental-work}

\subsection{Datasets}
\label{sec:datasets}

\begin{figure}[h!]
	\begin{center}
		\begin{subfigure}[t]{0.98\textwidth}
			\includegraphics[width=0.49\textwidth]{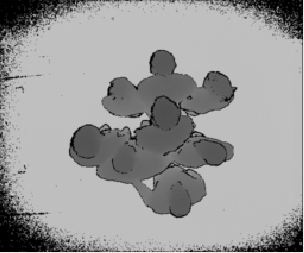}\;\includegraphics[width=0.49\textwidth]{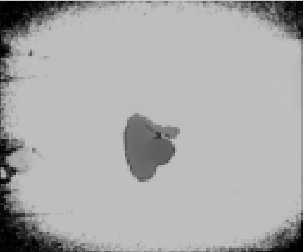}
			\caption{Sample frames from the \DBGOTPD{} dataset.}
			\label{fig:gotpd-samples}
		\end{subfigure}
		\begin{subfigure}[t]{0.98\textwidth}
			\includegraphics[width=0.49\textwidth]{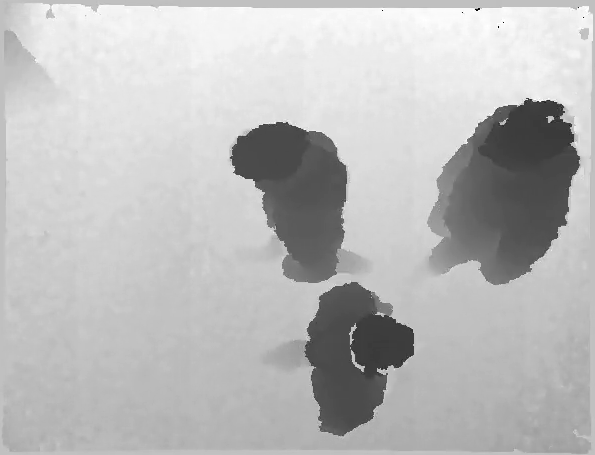}\;\includegraphics[width=0.49\textwidth]{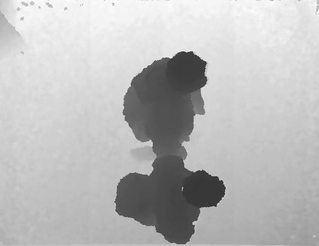}
			\caption{Sample frames from the \DBMIVIA{} dataset.}
			\label{fig:mivia-samples}
		\end{subfigure}
		\begin{subfigure}[t]{0.98\textwidth}
			\includegraphics[width=0.49\textwidth]{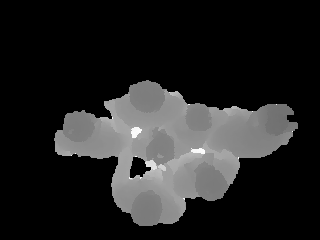}\;\includegraphics[width=0.49\textwidth]{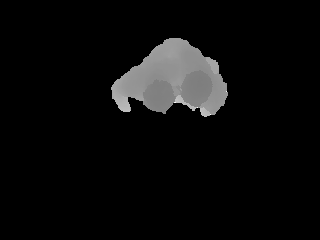}
			\caption{Sample frames from the \DBZHANG{} dataset.}
			\label{fig:zhang-samples}
		\end{subfigure}
	\end{center}
	\caption{Sample frames from the databases used (gray coded depth).}
	\label{fig:sample-frames}
\end{figure}

In order to provide a wide range of evaluation conditions, we have used
three different databases, that will be described
next. Figure~\ref{fig:sample-frames} provides some sample frames to give
an idea on the style and quality of the different datasets.

\subsubsection{GOTPD1 database}
\label{sec:gotpd1-database}

In this work, we have used part of the GOTPD1 database (available
at~\cite{gotpd1-2016} and \href{https://www.kaggle.com/lehomme/overhead-depth-images-people-detection-gotpd1}{GOTPD1}, that was recorded
with a Kinect\textsuperscript{\textregistered} v2 device located at a
height of $3.4m$. The recordings covered a broad variety of conditions,
with scenarios comprising single and multiple people, single and
multiple non-people (such as chairs), people with and without
accessories (hats, caps), people with different complexity, height, hair
color, and hair configurations, people actively moving and performing
additional actions (such as using their mobile phones, moving their
fists up and down, moving their arms, etc.).

The GOTPD1 data was originally split in two subsets, one for training
and the other for testing, with the configuration described
in~\cite{luna2017}. However, for this work, and given the strict
training data requirements of deep learning approaches, we will be using
the full GOTPD1 dataset for training purposes, and we recorded 23
additional sequences to allow for a proper evaluation effort. In this
manner, the training and testing subsets are fully independent, as
people and conditions differ between training and testing recordings.

Table~\ref{tab:gotpd1-training} and Table~\ref{tab:gotpd1-testing} show
the details of the training and testing subsets,
respectively. \textit{\#Frames} refers to the number of frames in the
recorded sequences, while \textit{\#Positives} refers to the number of
all the heads over all the frames (in our recordings we used 39
different people). The database contains sequences in which the users
were instructed on how to move under the camera (to allow for proper
coverage of the recording area), and sequences where people moved freely
(to allow for a more natural behavior)\footnote{This is fully detailed
	in the documentation distributed with the database
	at~\cite{gotpd1-2016} and \href{https://www.kaggle.com/lehomme/overhead-depth-images-people-detection-gotpd1}{GOTPD1}.}.

\begin{table*}[]
	\centering
	\caption{GOTPD1+ Training subset details.}
	\label{tab:gotpd1-training}
	\resizebox{\textwidth}{!}{
		\begin{tabular}{@{}cccc@{}}
			\hline
			\textbf{$\#Sequences$}                         & \textbf{$\#Frames$} & \textbf{$\#Positives$}    & \textbf{Description}      \\ \hline\hline
			\begin{tabular}[c]{@{}c@{}}S0001 through S0022,\\S0102,S0110,S0123,\\S0130,S0136,S0142\end{tabular}   & $26853$ & $14381$                    & Single people sequences   \\ \cline{1-4}
			\begin{tabular}[c]{@{}c@{}}S0023 through S0026\end{tabular}   & $7138$  & $2416$                 & Two people sequences \\ \cline{1-4}
			\begin{tabular}[c]{@{}c@{}}S0027 through S0029,\\S0031 through S0034,\\ S0152, S0153\end{tabular}   & $6526$  & $18106$                 & Multiple people sequences \\ \cline{1-4}
			\begin{tabular}[c]{@{}c@{}}S0035,\\S0038 through S0040\end{tabular}                           & $2323$ & $1458$      & \begin{tabular}[c]{@{}c@{}}Sequences with chairs and \\people
				balancing fists facing up\end{tabular} \\ \hline\hline
			\textbf{Totals}                               & $42840$ & $36361$             &  \\ \cline{1-3}
	\end{tabular}}
\end{table*}

\begin{table*}[]
	\centering
	\caption{GOTPD1+ testing subset details.}
	\label{tab:gotpd1-testing}
	\resizebox{\textwidth}{!}{    \begin{tabular}{@{}cccc@{}}
			\hline
			\textbf{Sequence IDs}                         & \textbf{$\#Samples$}  & \textbf{$\#Positives$}    & \textbf{Description}      \\ \hline\hline
			\begin{tabular}[c]{@{}c@{}}S041, S0101,\\S0131 through S0135, S143\end{tabular}        & $2742$ & $2391$                   & Single people sequences   \\ \cline{1-4}
			S0144 through S0146                           & $1888$ & $2865$                    & Two people sequences \\ \cline{1-4}
			\begin{tabular}[c]{@{}c@{}}S0030, S0147 through S0151,\\S0154 through 157\end{tabular} & $4516$ & $14837$                   & Multiple people sequences \\ \cline{1-4}
			S0036 through S0037                           & $1789$ & $0$                       & Sequences with chairs \\ \hline\hline
			\textbf{Totals}                               & $11375$ & $21403$             &  \\ \cline{1-3}
	\end{tabular}}
\end{table*}

This dataset (composed by the original GOTPD1 and the additional 23
recorded sequences), will be referred to as \DBGOTPD{} in the tables
included in Section~\ref{sec:results-discussion}. The first row of
figure~\ref{fig:sample-frames} shows two sample frames from this
dataset.

\subsubsection{MIVIA People Counting Dataset}
\label{sec:del-pizzo-2016}

The MIVIA people counting dataset (available at~\cite{mivia2016}) has been
developed by the MIVIA Research Lab in the University of Palermo and was
first described and used in~\cite{delpizzo2016counting}. It is mainly
aimed at evaluating people flow density in the sensed environment, but
we will use it for people counting and tracking purposes.

The MIVIA dataset includes RGB and depth data captured with a
Kinect\textsuperscript{\textregistered} device located in an overhead
position at a height of $3.20m$. The recordings have been captured in
indoor and outdoor conditions, and for isolated and group
transits. From the 34 available sequences, we will be using 8 of them,
all that correspond to indoor conditions for both the single people and
multiple people transits. The recordings include people walking with
bags and other big objects, and their details are shown in
Figure~\ref{tab:mivia-testing}.

\begin{table*}[]
	\centering
	\caption{MIVIA testing subset details.}
	\label{tab:mivia-testing}
	\resizebox{\textwidth}{!}{    \begin{tabular}{@{}cccc@{}}
			\hline
			\textbf{Sequence IDs}   & \textbf{$\#Frames$}  & \textbf{$\#Positives$} & \textbf{Description}      \\ \hline\hline
			DIS1 DIS2               & 7692                    & 3601                   & Single people sequences   \\ \cline{1-4}
			DIS3 DIS4 DIS5          & 10259                   & 5315                   & Two people sequences      \\ \cline{1-4}
			DIG1 DIG2 DIG3          & 12222                   & 8780                   & Multiple people sequences \\ \hline\hline
			\textbf{Totals}         & 30173                   & 17696                  &                           \\ \cline{1-3}
	\end{tabular}}
\end{table*}

In our experiments we will use the depth streams with a resolution of
${640\times480}$. Because of the different sizes, it will be necessary to
resize the depth streams used in this database to the sizes that
\DPDNet{} can use.

This dataset will be referred to as \DBMIVIA{} in the tables included in
Section~\ref{sec:results-discussion}. Middle row of
figure~\ref{fig:sample-frames} shows two sample frames from this
dataset.

\subsubsection{Zhang 2012 database}
\label{sec:zhang-2012-database}

The Zhang 2012 database~\cite{Zhang2012}
consists of two sequences (\texttt{dataset1} and \texttt{dataset2}) recorded with a
Kinect\textsuperscript{\textregistered} device also located in an
overhead position, generating depth data with a resolution of $320\times240$
pixels. Both sequences include multiple people moving under the camera,
and their details are shown in Table~\ref{tab:zhang2012-testing}. The
recordings include groups of people walking closely to each other,
people walking freely and people with bags and other big objects.

The data, kindly provided to us by its authors, is processed with an
ad-hoc background subtraction strategy (described in~\cite{Zhang2012}),
that induces some artifacts in the depth
images. \begin{table*}[]
	\centering
	\caption{Zhang2012 testing subset details.}
	\label{tab:zhang2012-testing}
	\resizebox{\textwidth}{!}{    \begin{tabular}{@{}cccc@{}}
			\hline
			\textbf{Sequence IDs}  & \textbf{$\#Positives$}  & \textbf{$\#Positives$}  & \textbf{Description}        \\ \hline\hline
			dataset1               & 2384                    & 4527                    & Multiple people sequences   \\ \cline{1-4}
			dataset2               & 1500                    & 1553                    & Multiple people sequences   \\ \hline\hline
			\textbf{Totals}        & 3884                    & 6080                    &                             \\ \cline{1-3}
	\end{tabular}}
\end{table*}

This dataset will be referred to as \DBZHANG{} in the tables included in
section~\ref{sec:results-discussion}. Last row of
figure~\ref{fig:sample-frames} shows two sample frames from this
dataset.
\subsection{Experimental setup}
\label{sec:experimental-setup}

In all the experiments carried out, we use the three different datasets
described in section~\ref{sec:datasets}.

We compare the performance of our proposals \DPDNet{} and
\DPDNetFast{} with other strategies described in the literature that use
an overhead ToF camera in a people detection/tracking/counting task. The following 
state-of-the-art methods are used for comparison: 
\begin{itemize}
	\item \WaterFilling{}~\cite{Zhang2012}: this method is based on the
	\emph{water filling} algorithm and the original source code was kindly
	provided to us by its authors.  To run this method in the \DBGOTPD{}
	dataset we scaled the images to $640\times480$ pixels.
	\item \SLICEPCA{}~\cite{luna2017}: this method uses feature vectors
	extracted from the depth image and a generative PCA model of these
	vectors to classify detections.  In our experiments, the PCA model is
	trained using the GOTPD1 dataset. In this regard, we train two
	categories: people and non-people.
	\item \SLICESVM{}~\cite{fernandez2017}: this method is similar to
	\SLICEPCA{}, but uses a SVM classifier instead of PCA. In our
	experiments, the SVM model is trained again using the full GOTPD1
	dataset.  \item
	\MexicanHat{}~\cite{Stahlschmidt2013,Stahlschmidt2014}: this method is
	based on the normalized Mexican Hat Wavelet and requires the detection
	and removal of the floor plane. It does not require training data
	though.
\end{itemize}

It is important to note that in the comparison, we did not use a
tracking module for any of the methods. Our objective is to provide a
fair comparison on their discrimination capabilities without other
improvements. The only input frame manipulation we did was that required
to scale the frame size to fit in the input stage conditions of the
evaluated algorithms.

\subsection{Evaluation Metrics}
\label{sec:evaluation-metrics}

To provide a detailed view on the performance on the evaluated
algorithms, we will calculate the main standard metrics in a detection
problem, namely: $Precision$, $Recall$, $F1_{score}$, False Negative
Rate ($FNR$), False Positive Rate ($FPR$), and overall error $ERR$
(calculated as $ERR=FNR+FPR$)

We will also calculate confidence intervals for the $F1_{score}$ and
$ERR$ metrics, for a confidence value of $95\%$, to assess the
statistical significance of the results when comparing different
strategies.

\subsection{Results and Discussion}
\label{sec:results-discussion}

In this section we will first provide detailed results for all the
algorithms described in Section~\ref{sec:experimental-setup} applied to
each of the datasets described in Section~\ref{sec:datasets}.

Tables~\ref{tab:results-uah-all}, \ref{tab:results-ita-all}
and~\ref{tab:results-chi-all} include the results for the
\DBGOTPD{},\DBMIVIA{} and \DBZHANG{} datasets respectively, including
the evaluation metrics described in section~\ref{sec:evaluation-metrics}
for all the evaluated algorithms. To visually aid in the analysis of the
results, we have added a green background to those table cells that
correspond to the best results across all algorithms for each condition.
\begin{table*}[!htbp]
	\centering
	\caption{Detailed Results using all the tested algorithms on the
		\DBGOTPD{} dataset. All values in $\%$ (cells in green background
		correspond to the best results across all algorithms for each
		condition).                          }
	\label{tab:results-uah-all}
	\resizebox{\textwidth}{!}{    \begin{tabular}{cccccccc}       & DB       & $Precision$ & $Recall$ & $F1_{score}$  & $FNR$ & $FPR$ & $ERR$ \\\hline\hline
			\multirow{5}{*}{\MexicanHat{}}      
			& Single person        & $99.87$ & $99.34$ & $99.60 \pm 0.23$ & $0.66$  & $0.16$  &  $0.44 \pm 0.25$ \\\cline{2-8}
			& Two people           & $99.63$ & $49.72$ & $66.34 \pm 1.91$ & $50.28$ & $0.41$  & $34.84 \pm 1.93$ \\\cline{2-8}
			& More than two people & $99.86$ & $37.84$ & $54.88 \pm 0.87$ & $62.16$ & $0.47$  & $55.92 \pm 0.87$ \\\cline{2-8}
			& Chairs. no people    &      &      &               &      & $21.97$ & $21.97 \pm 1.92$ \\\cline{2-8}
			& Totals               & $94.24$ & $45.63$ & $61.48 \pm 0.68$ & $54.37$ & $8.06$  & $42.48 \pm 0.69$ \\\hline\hline
			\multirow{5}{*}{\SLICESVM{}}      
			& Single person*       & $98.23$ & \cellcolor{111} $\mbf{99.87}$ & $99.04 \pm 0.36$ & \cellcolor{111} $\mbf{0.13}$ & $2.18$ & $1.06 \pm 0.38$ \\\cline{2-8}
			& Two people*          & $93.17$ & \cellcolor{111} $\mbf{100.00}$ & $96.46 \pm 0.75$ & \cellcolor{111} $\mbf{0.00}$ & $16.25$ & $5.05 \pm 0.89$ \\\cline{2-8}
			& More than two people*& $99.93$ & $99.26$ & $99.59 \pm 0.11$ & $0.74$ & $0.63$ & $0.73 \pm 0.15$ \\\cline{2-8}
			& Chairs. no people*   &  &  &  &  & $35.33$ & $35.33 \pm 2.21$ \\\cline{2-8}
			& Totals*              & $96.71$ & $99.40$ & $98.04 \pm 0.19$ & $0.60$ & $9.70$ & $2.95 \pm 0.24$ \\\hline\hline
			\multirow{5}{*}{\SLICEPCA{}}           
			& Single person        & $98.23$ & \cellcolor{111} $\mbf{99.87}$ & $99.04 \pm 0.36$ & \cellcolor{111} $\mbf{0.13}$ &  $2.18$ & $1.06 \pm 0.38$ \\\cline{2-8}
			& Two people           & $94.10$ & \cellcolor{111} $\mbf{100.00}$ & $96.96 \pm 0.70$ & \cellcolor{111} $\mbf{0.00}$ & $13.91$ & $4.32 \pm 0.82$ \\\cline{2-8}
			& More than two people & $99.93$ &  $99.20$ & $99.56 \pm 0.12$ & $0.80$ &  $0.63$ & $0.79 \pm 0.15$ \\\cline{2-8}
			& Chairs. no people    &      &       &               &     &  $0.00$ & \cellcolor{111} $\mbf{0.00} \pm 0.00$ \\\cline{2-8}
			& Totals               & $99.06$ & $99.36$  & $99.21 \pm 0.12$ & $0.64$ &  $2.70$ & $1.18 \pm 0.15$ \\\hline\hline
			\multirow{5}{*}{\WaterFilling{}}      
			& Single person        & $98.18$ & $98.63$  & $98.40 \pm 0.47$ & $1.37$ &  $2.31$ &  $1.79 \pm 0.50$ \\\cline{2-8}
			& Two people           & $99.26$ & \cellcolor{111} $\mbf{100.00}$ & $99.63 \pm 0.25$ & \cellcolor{111} $\mbf{0.00}$ &  $1.66$ &  $0.51 \pm 0.29$ \\\cline{2-8}
			& More than two people & $99.90$ & $99.81$  & $99.85 \pm 0.07$ & $0.19$ &  $0.86$ &  $0.26 \pm 0.09$ \\\cline{2-8}
			& Chairs. no people	   &      &       &               &     & $22.97$ & $22.97 \pm 1.95$ \\\cline{2-8}
			& Totals               & $96.90$ & $99.70$  & $98.28 \pm 0.18$ & $0.30$ &  $9.24$ &  $2.59 \pm 0.22$ \\\hline\hline
			\multirow{5}{*}{\DPDNetFast{}}        
			& Single person        & \cellcolor{222} $\mbf{100.00}$ & $99.67$  & $99.84 \pm 0.15$ & $0.33$ & \cellcolor{222} $\mbf{0.00}$ & $0.18 \pm 0.16$ \\\cline{2-8}
			& Two people           & $99.88$  & \cellcolor{222} $\mbf{100.00}$ & $99.94 \pm 0.10$ & \cellcolor{222} $\mbf{0.00}$ & $0.28$ & $0.09 \pm 0.12$ \\\cline{2-8}
			& More than two people & $99.92$  & \cellcolor{222} $\mbf{99.90}$  & \cellcolor{222} $\mbf{99.91} \pm 0.05$ & $0.10$ & $0.71$ & \cellcolor{222} $\mbf{0.16} \pm 0.07$ \\\cline{2-8}
			& Chairs. no people    &       &       &               &     & $0.39$ & $0.39 \pm 0.29$ \\\cline{2-8}
			& Totals               & $99.88$  & \cellcolor{222} $\mbf{99.89}$  & \cellcolor{222} $\mbf{99.88} \pm 0.05$ & \cellcolor{222} $\mbf{0.11}$ & $0.36$ & \cellcolor{222} $\mbf{0.17} \pm 0.06$ \\\hline\hline
			\multirow{5}{*}{\DPDNet{}}        
			& Single person	         &  $99.93$ & \cellcolor{222} $\mbf{99.87}$ & \cellcolor{222} $\mbf{99.90} \pm 0.12$ & \cellcolor{222} $\mbf{0.13}$ & $0.08$ & \cellcolor{222} $\mbf{0.11} \pm 0.12$ \\\cline{2-8}
			& Two people	           & \cellcolor{222} $\mbf{100.00}$ & $99.94$ & \cellcolor{222} $\mbf{99.97} \pm 0.07$ & $0.06$ & \cellcolor{222} $\mbf{0.00}$ & \cellcolor{222} $\mbf{0.04} \pm 0.08$ \\\cline{2-8}
			& More than two people	 & \cellcolor{222} $\mbf{100.00}$ & $99.71$ & $99.86 \pm 0.07$ & $0.29$ & \cellcolor{222} $\mbf{0.00}$ & $0.26 \pm 0.09$ \\\cline{2-8}
			& Chairs. no people	     &     &    &    &   & \cellcolor{222} $\mbf{0.00}$ & \cellcolor{222} $\mbf{0.00} \pm 0.00$ \\\cline{2-8}
			& Totals	               &  \cellcolor{222} $\mbf{99.99}$ & $99.75$ & $99.87 \pm 0.05$ & $0.25$ & \cellcolor{222} $\mbf{0.02}$ & $0.19 \pm 0.06$ \\\hline
	\end{tabular}}
\end{table*}

\subsubsection{Results for the \DBGOTPD{} database}
\label{sec:results-dbgotpd-data}

Regarding the results for the \DBGOTPD{} database
(Table~\ref{tab:results-uah-all}), it is clearly seen that the \DPDNet{}
and \DPDNetFast{} solutions are the best ones, and this is specially
significant for sequences including more than two people. Despite this,
\SLICEPCA{} and \SLICESVM{} also obtain top results for the $Recall$ and
$FNR$ rates, for the sequences with one and two people. This happens at
the cost of significantly increase the $FPR$ rates, which is the reason
why these methods finally got worse performance in the $F1_{score}$ and
$ERR$ metrics. The \WaterFilling{} algorithm has also top performance
for the sequences with two people, but also with higher $FPR$ rates. The
\MexicanHat{} algorithm is clearly the worst one, as has been already
proved in previous works.

Despite the good results of \SLICESVM{} and \WaterFilling{}, their
robustness to sequences without people (just chairs) is less than that
of \DPDNet{}, \SLICEPCA{} and \DPDNetFast{}. In the case of \SLICESVM{},
this bad performance could be due to the class modeling carried out, and
for the \WaterFilling{} algorithm, it is clearly due to the absence of
an explicit discrimination procedure, which generates a higher number of
false positives.  The results obtained with the \SLICEPCA{} method in
the chair sequences are similar or better than those obtained with our
CNN based proposals, but we must bear in mind that the \SLICEPCA{}
method requires a calibration process to know the intrinsic and
extrinsic parameters of the sensor.

It is also worth mentioning that the DNN based methods are undoubtedly
the best when considering the $FPR$ rates, showing a very good
discrimination capability.

Finally, when comparing \DPDNet{} and \DPDNetFast{}, a singular effect can be
observed: the $FNR$ values for \DPDNetFast{} are better than those of
\DPDNet{}, while the $FPR$ values of \DPDNet{} are better than those of
\DPDNetFast{}. This can be explained by the reduction in the image
resolution, since this reduction implies an interpolation that allows to
slightly filter the noise and artifacts in the input image. This can
lead the fast version to have an easier job in discriminating negatives,
while in the case of the conventional version, the availability of the
full image information allows for a better discrimination of actual
people present in the scene.

\begin{table*}[!htbp]
	\centering
	\caption{Detailed Results using all the tested algorithms on the
		\DBMIVIA{} dataset. All values in $\%$ (cells in green background
		correspond to the best results across all algorithms for each condition).}
	\label{tab:results-ita-all}
	\resizebox{\textwidth}{!}{    \begin{tabular}{cccccccc}       & DB       & $Precision$ & $Recall$ & $F1_{score}$  & $FNR$ & $FPR$ & $ERR$ \\\hline\hline
			\multirow{4}{*}{\MexicanHat{}}      
			& Single person        & $93.76$ & $92.66$ & $93.21 \pm 0.56$ & $7.34$ & $2.94$ & $4.36 \pm 0.46$\\\cline{2-8}
			& Two people           & $88.61$ & $86.64$ & $87.61 \pm 0.64$ & $13.36$ & $6.21$ & $8.77 \pm 0.55$\\\cline{2-8}
			& More than two people & $93.73$ & $53.67$ & $68.26 \pm 0.76$ & $46.33$ & $2.65$ & $21.21 \pm 0.67$\\\cline{2-8}
			& Totals               & $91.81$ & $71.40$ & $80.33 \pm 0.43$ & $28.60$ & $3.89$ & $13.27 \pm 0.37$\\\hline\hline
			\multirow{4}{*}{\SLICESVM{}}      
			& Single person*       & $92.92$ & $93.90$ & $93.41 \pm 0.55$ & $6.10$ & $3.39$ & $4.26 \pm 0.45$\\\cline{2-8}
			& Two people*          & $87.75$ & $87.97$ & $87.86 \pm 0.63$ & $12.03$ & $6.85$ & $8.70 \pm 0.55$\\\cline{2-8}
			& More than two people*& $91.85$ & $94.11$ & $92.96 \pm 0.42$ & $5.89$ & $6.13$ & $6.03 \pm 0.39$\\\cline{2-8}
			& Totals*              & $90.85$ & $92.22$ & $91.53 \pm 0.30$ & $7.78$ & $5.66$ & $6.46 \pm 0.27$\\\hline\hline
			\multirow{4}{*}{\SLICEPCA{}}           
			& Single person*       & $93.30$ & $93.90$ & $93.67 \pm 0.55$ & $6.10$ & $3.20$ & $4.13 \pm 0.44$\\\cline{2-8}
			& Two people*          & $87.97$ & $87.97$ & $87.97 \pm 0.63$ & $12.03$ & $6.71$ & $8.62 \pm 0.54$\\\cline{2-8}
			& More than two people*& $92.91$ & $93.95$ & $93.43 \pm 0.40$ & $6.05$ & $5.25$ & $5.59 \pm 0.38$\\\cline{2-8}
			& Totals*              & $91.52$ & $92.14$ & $91.83 \pm 0.30$ & $7.86$ & $5.20$ & $6.20 \pm 0.26$\\\hline\hline
			\multirow{4}{*}{\WaterFilling{}}      
			& Single person        & $95.20$ & \cellcolor{222}$\mbf{98.96}$ & $97.04 \pm 0.38$ & \cellcolor{222}$\mbf{1.04}$ & $2.41$ & $1.96 \pm 0.31$\\\cline{2-8}
			& Two people           & $96.67$ & \cellcolor{222}$\mbf{99.84}$ & $98.23 \pm 0.26$ & \cellcolor{222}$\mbf{0.16}$ & $1.96$ & $1.31 \pm 0.22$\\\cline{2-8}
			& More than two people & $91.29$ & \cellcolor{222}$\mbf{98.53}$ & $94.77 \pm 0.37$ & \cellcolor{222}$\mbf{1.47}$ & $6.81$ & $4.57 \pm 0.34$\\\cline{2-8}
			& Totals               & $93.68$ & \cellcolor{222}$\mbf{99.02}$ & $96.27 \pm 0.21$ & \cellcolor{222}$\mbf{0.98}$ & $4.08$ & $2.91 \pm 0.18$\\\hline\hline
			\multirow{4}{*}{\DPDNetFast{}}        
			& Single person        & $99.71$ & $97.41$ & $98.55 \pm 0.27$ & $2.59$ & $0.14$ & $0.94 \pm 0.22$\\\cline{2-8}
			& Two people           & $99.54$ & $99.09$ & $99.32 \pm 0.16$ & $0.91$ & $0.26$ & $0.50 \pm 0.14$\\\cline{2-8}
			& More than two people & \cellcolor{222}$\mbf{99.24}$ & $96.62$ & $97.91 \pm 0.23$ & $3.38$ & \cellcolor{222}$\mbf{0.53}$ & $1.72 \pm 0.21$\\\cline{2-8}
			& Totals               & $99.43$ & $97.54$ & $98.47 \pm 0.13$ & $2.46$ & $0.34$ & $1.14 \pm 0.12$\\\hline\hline
			\multirow{4}{*}{\DPDNet{}}        
			& Single person        & \cellcolor{222}$\mbf{100.00}$ & $98.56$ & \cellcolor{222}$\mbf{99.27} \pm 0.19$ & $1.44$ & \cellcolor{222}$\mbf{0.00}$ & \cellcolor{222}$\mbf{0.47} \pm 0.15$\\\cline{2-8}
			& Two people           & \cellcolor{222}$\mbf{99.81}$ & $99.05$ & \cellcolor{222}$\mbf{99.43} \pm 0.15$ & $0.95$ & \cellcolor{222}$\mbf{0.11}$ & \cellcolor{222}$\mbf{0.41} \pm 0.12$\\\cline{2-8}
			& More than two people & $99.23$ & $98.14$ & \cellcolor{222}$\mbf{98.68} \pm 0.19$ & $1.86$ & $0.54$ & $1.09 \pm 0.17$\\\cline{2-8}
			& Totals               & \cellcolor{222}$\mbf{99.57}$ & $98.50$ & \cellcolor{222}$\mbf{99.03} \pm 0.11$ & $1.50$ & \cellcolor{222}$\mbf{0.26}$ & \cellcolor{222}$\mbf{0.72} \pm 0.09$\\\hline
	\end{tabular}}
\end{table*}

\subsubsection{Results for the \DBMIVIA{} database}
\label{sec:results-dbmivia-data}

Regarding the results for the \DBMIVIA{} database
(Table~\ref{tab:results-ita-all}), we first have to stress and remind
the fact that no training nor adaptation has been done for this new
dataset.

Again, the best results are found in the CNN-based methods (which imply
the use of a model trained on a different dataset), and (in specific
metrics) in the \WaterFilling{} algorithm (that does not need any
training procedure).

The \SLICEPCA{} and \SLICESVM{} methods exhibit a relevant performance
drop that can be explained by their higher sensitivity to the mismatch
in the recording conditions as compared with those of the training
material used.

The better performance of the \WaterFilling{} algorithm in the $Recall$
and $FNR$ rates is again achieved at the cost of increasing the $FPR$
rates, which finally translates to overall lower $F1_{score}$ and $ERR$
metrics.

As in the case of \DBGOTPD{}, the \MexicanHat{} method is the worst of
all in all metrics.

\begin{table*}[!htbp]
	\centering
	\caption{Detailed Results using all the tested algorithms on the
		\DBZHANG{} dataset. All values in $\%$ (cells in green background
		correspond to the best results across all algorithms for each condition).}
	\label{tab:results-chi-all}
	\resizebox{\textwidth}{!}{    \begin{tabular}{cccccccc}        & DB       & $Precision$ & $Recall$ & $F1_{score}$  & $FNR$ & $FPR$ & $ERR$ \\\hline
			\multirow{3}{*}{\MexicanHat{}}      
			& Dataset1 & $40.52$ & $24.51$ & $30.54 \pm 1.51$ & $75.49$ & $410.76$ & $102.49 \pm -$\\\cline{2-8}
			& Dataset2 & $26.72$ & $21.02$ & $23.53 \pm 2.02$ & $78.98$ & $86.32$ & $81.92 \pm 1.83$\\\cline{2-8}
			& Totals   & $36.56$ & $23.68$ & $28.74 \pm 1.22$ & $76.32$ & $182.85$ & $95.87 \pm 0.54$\\\hline
			\multirow{3}{*}{\SLICESVM{}}      
			& Dataset1*& $82.92$ & $85.77$ & $84.32 \pm 1.19$ & $14.32$ & $226.82$ & $29.60 \pm 1.49$\\\cline{2-8}
			& Dataset2*& $59.56$ & $59.28$ & $59.42 \pm 2.33$ & $40.72$ & $64.38$  & $49.82 \pm 2.37$\\\cline{2-8}
			& Totals*  & $77.50$ & $79.44$ & $78.46 \pm 1.10$ & $20.56$ & $110.57$ & $36.09 \pm 1.29$\\\hline
			\multirow{3}{*}{\SLICEPCA{}}           
			& Dataset1*& $94.31$ & $97.81$ & $96.03 \pm 0.64$ & $2.19$ & $67.60$ & $7.44 \pm 0.86$\\\cline{2-8}
			& Dataset2*& $94.48$ & $99.30$ & $96.83 \pm 0.83$ & $0.70$ & $8.36$ & $3.84 \pm 0.92$\\\cline{2-8}
			& Totals*  & $94.35$ & $98.16$ & $96.22 \pm 0.52$ & $1.84$ & $25.69$ & $6.28 \pm 0.66$\\\hline
			\multirow{3}{*}{\WaterFilling{}}      
			& Dataset1 & $95.40$ & $98.76$ & $97.05 \pm 0.55$ & $1.24$ & $55.48$ & $5.53 \pm 0.75$\\\cline{2-8}
			& Dataset2 & $96.70$ & $98.91$ & $97.79 \pm 0.70$ & $1.09$ & $4.96$ & $2.66 \pm 0.77$\\\cline{2-8}
			& Totals   & $95.70$ & $98.79$ & $97.22 \pm 0.44$ & $1.21$ & $19.71$ & $4.61 \pm 0.57$\\\hline
			\multirow{3}{*}{\DPDNetFast{}}        
			& Dataset1 & \cellcolor{222}$\mbf{99.57}$ & $99.21$ & \cellcolor{222}$\mbf{99.39} \pm 0.25$ & $0.79$ & \cellcolor{222}$\mbf{4.95}$ & \cellcolor{222}$\mbf{1.12} \pm 0.34$\\\cline{2-8}
			& Dataset2 & \cellcolor{222}$\mbf{99.01}$ & $99.11$ & \cellcolor{222}$\mbf{99.06} \pm 0.46$ & $0.89$ & \cellcolor{222}$\mbf{1.45}$ & \cellcolor{222}$\mbf{1.12} \pm 0.50$\\\cline{2-8}
			& Totals   & \cellcolor{222}$\mbf{99.44}$ & $99.19$ & \cellcolor{222}$\mbf{99.31} \pm 0.22$ & $0.81$ & \cellcolor{222}$\mbf{2.47}$ & \cellcolor{222}$\mbf{1.12} \pm 0.28$\\\hline
			\multirow{3}{*}{\DPDNet{}}        
			& Dataset1 & $98.89$ & \cellcolor{222}$\mbf{99.76}$ & $99.32 \pm 0.27$ & \cellcolor{222}$\mbf{0.24}$ & $12.94$ & $1.26 \pm 0.36$\\\cline{2-8}
			& Dataset2 & $97.10$ & \cellcolor{222}$\mbf{99.90}$ & $98.48 \pm 0.58$ & \cellcolor{222}$\mbf{0.10}$ & $4.35$ & $1.83 \pm 0.64$\\\cline{2-8}
			& Totals   & $98.46$ & \cellcolor{222}$\mbf{99.79}$ & $99.12 \pm 0.25$ & \cellcolor{222}$\mbf{0.21}$ & $6.87$ & $1.44 \pm 0.32$\\\hline
	\end{tabular}}
\end{table*}

\subsubsection{Results for the \DBZHANG{} database}
\label{sec:results-dbzh-datab}

Regarding the results for the \DBZHANG{} database
(Table~\ref{tab:results-chi-all}), we are again in the case of a
mismatch between the training and evaluation conditions discussed above.

Here, the superiority of the CNN-based methods are even clearer than in
the two previous datasets, obtaining very good results considering the
experimental conditions.

The performance drop of the \SLICEPCA{} and \SLICESVM{} methods is
higher than in the \DBMIVIA{} results (specially in the latter). As
mentioned in Section~\ref{sec:zhang-2012-database}, images of this
dataset contain artifacts from the background subtraction method, and
this might affect the classifiers \SLICEPCA{} and \SLICESVM{}, which
were trained with cleaner data.

As discussed in the case of the \DBMIVIA{} results, the \WaterFilling{}
algorithm performance is closer to the top results in the $Recall$ and
$FNR$ rates, with the same explanation than above.

The CNN-based methods are affected by image artifacts to a much lower
extent than \SLICEPCA{} and
\SLICESVM{}.              Between  the two proposals, \DPDNet{} is more affected than
\DPDNetFast{}, possibly given that its greater number of parameters (due
to its the larger input image size) forces it to search for more defined
and detailed features similar to those found in the training subset of
the \DBGOTPD{} dataset, while \DPDNetFast{} generalizes better when
facing new datasets, due to the lower definition of its input images
(finding more general characteristics of people).

Finally, as in all the previous cases, the \MexicanHat{} method gets the
worst performance for all metrics.

\subsubsection{Overall comparison}
\label{sec:overal-comparison}

Once the detailed results have been discussed for each dataset, now we
will provide full details on the overall comparison among the evaluated
algorithms for all the datasets used. We will include tables with the
overall behavior of the different strategies, and we will also include
bar charts to ease the visual comparison, providing error confidence
values for all the metrics. In the tables and graphics below, we have
omitted the results for the \MexicanHat{} algorithm due to its bad
performance. In the charts, please note that the vertical scale is not
the same for all of them, as they have been adjusted to allow for a
better visualization.

From the previous discussion on the results show in tables
\ref{tab:results-uah-all}, \ref{tab:results-ita-all} and
\ref{tab:results-chi-all}, it can be concluded that the proposed
solutions \DPDNet{} and \DPDNetFast{} are the best alternatives for a
robust detection in this application. They have obtained the best
results, and they have proved to be robust against changing database
recording conditions, maintaining the overall error rate around $1\%$
percent in all cases. The classic trainable methods such as \SLICEPCA{}
and \SLICESVM{} are good alternatives but have worse results, and they
require calibration and retraining to change the detection environment
to provide optimal results. On the other hand, \WaterFilling{} can be
seen as another alternative system, but unlike the two previous
approaches, it acts more as a detector of maximums than as a
classification algorithm. Finally, the \MexicanHat{} algorithm is not a
good alternative system and globally has obtained the worst results
among the evaluated systems.

\begin{figure}[h]
	\begin{center}
		\begin{subfigure}[t]{0.49\textwidth}
			\includegraphics[width=\linewidth]{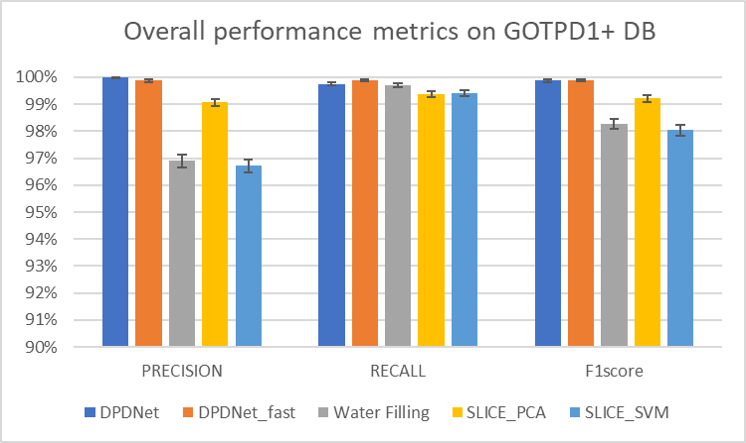}
			\caption{Performance metrics.}
			\label{fig:gotpd1+-performance}.
		\end{subfigure}
		\;    \begin{subfigure}[t]{0.49\textwidth}
			\includegraphics[width=\linewidth]{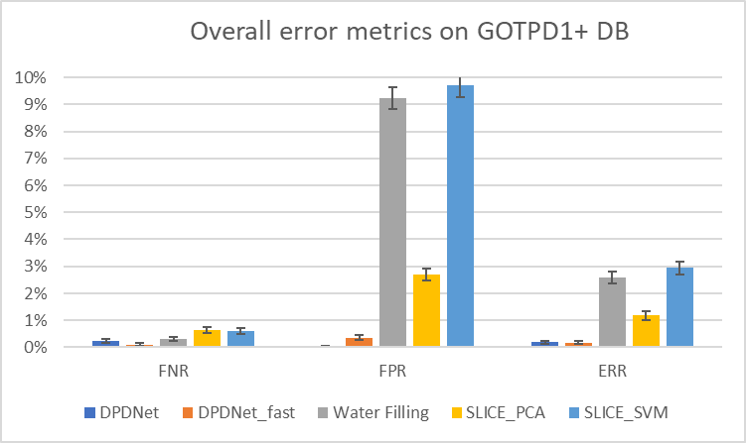}
			\caption{Error metrics.}
			\label{fig:gotpd1+-error}.
		\end{subfigure}
	\end{center}
	\caption{Overall performance and error metrics for the \DBGOTPD{} dataset.}
	\label{fig:results-gotpd1+}
\end{figure}

Table~\ref{tab:results-gotpd1+} shows the final aggregated results of
the evaluated algorithms on the \texttt{GOTPD1+} dataset, and in
Figures~\ref{fig:gotpd1+-performance} and~\ref{fig:gotpd1+-error} we
show the graphical comparison among the different strategies.

\begin{table*}[h]
	\centering
	\caption{Results for the \DBGOTPD{} dataset. All values in $\%$.}
	\label{tab:results-gotpd1+}
	\resizebox{\textwidth}{!}{    \begin{tabular}{c|ccc|ccc}
			\hline    \hline
			DB            & $Precision$       &   $Recall$       & $F_1score$             &  $FNR$        &  $FPR$       & $ERR$              \\\hline\hline
			\DPDNet       & \cellcolor{222}$\mbf{99.99}$     & $99.75$          & $99.87       \pm 0.05$ & $0.25$        & \cellcolor{222}$\mbf{0.02}$ &        $0.19 \pm 0.06$  \\\hline
			\DPDNetFast   & $99.88$           & \cellcolor{222}$\mbf{99.89}$    & \cellcolor{222}$\mbf{99.88} \pm 0.05$ & \cellcolor{222}$\mbf{0.11}$  & $0.36$       &   \cellcolor{222}$\mbf{0.17 \pm 0.06}$  \\\hline
			\WaterFilling & $96.90$           & $99.70$          & $98.28       \pm 0.18$ & $0.30$        & $9.24$       &        $2.59 \pm 0.22$  \\\hline
			\SLICEPCA{}*  & $99.06$           & $99.36$          & $99.21       \pm 0.12$ & $0.64$        & $2.70$       &        $1.18 \pm 0.15$  \\\hline
			\SLICESVM{}*  & $96.71$           & $99.40$          & $98.04       \pm 0.19$ & $0.60$        & $9.70$      &        $2.95 \pm 0.24$  \\\hline
			\MexicanHat   & $94.24$           & $45.63$          & $61.48       \pm 0.68$ & $54.37$       & $8.06$       &       $42.48 \pm 0.69$  \\\hline    \hline
	\end{tabular}}
\end{table*}

Figure~\ref{fig:gotpd1+-performance} shows how the performance metrics
are reasonably high, except for the \WaterFilling{} and \SLICESVM{}
algorithms, whose $Precision$ and $F1_{score}$ have a significant
drop. This tendency is also clearly observed in the error metrics of
Figure~\ref{fig:gotpd1+-error}. The CNN-based methods achieve the best
results, and the observed improvements are statistically significant
against the non CNN-based methods, considering the included confidence
bands.

\begin{figure}[h]
	\begin{center}
		\begin{subfigure}[t]{0.49\textwidth}
			\includegraphics[width=\linewidth]{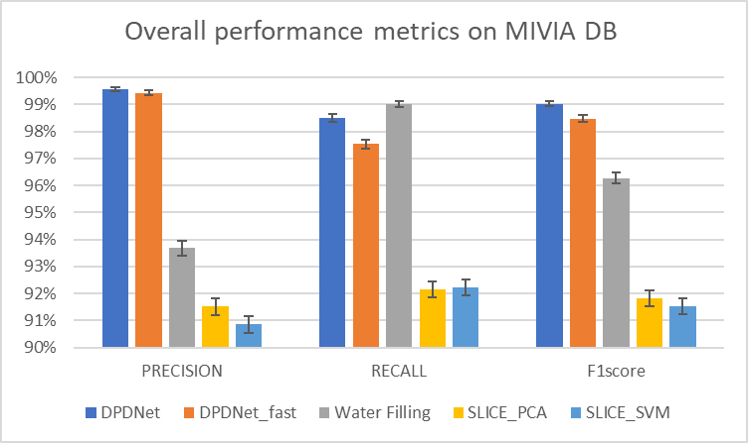}
			\caption{Performance metrics.}
			\label{fig:mivia-performance}.
		\end{subfigure}
		\;    \begin{subfigure}[t]{0.49\textwidth}
			\includegraphics[width=\linewidth]{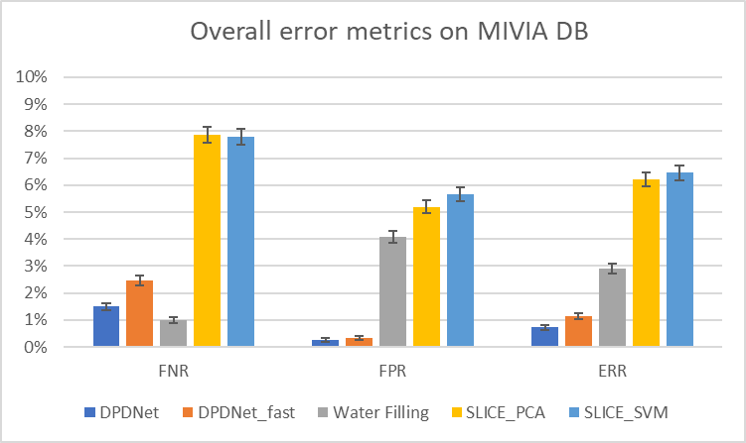}
			\caption{Error metrics.}
			\label{fig:mivia-error}.
		\end{subfigure}
	\end{center}
	\caption{Overall performance and error metrics for the \DBMIVIA{} dataset.}
	\label{fig:results-mivia}
\end{figure}

Table~\ref{tab:results-mivia} shows the final aggregated results of the evaluated
algorithms on the \texttt{MIVIA} dataset, and in
Figures~\ref{fig:mivia-performance} and~\ref{fig:mivia-error} we show
the graphical comparison among the different strategies

\begin{table*}[h]
	\centering
	\caption{Results for the \DBMIVIA{} dataset. All values in $\%$.}
	\label{tab:results-mivia}
	\resizebox{\textwidth}{!}{    \begin{tabular}{c|ccc|ccc}
			\hline
			DB            & $Precision$     &  $Recall$     &	$F_1score$             &    $FNR$  &  $FPR$    & $ERR$                  \\\hline\hline
			\DPDNet       & \cellcolor{222}$\mbf{99.57}$   & \cellcolor{222}$\mbf{98.50}$ & \cellcolor{222}$\mbf{99.03 \pm 0.11}$ &  \cellcolor{222}$\mbf{1.50}$ &  \cellcolor{222}$\mbf{0.26}$ &   \cellcolor{222}$\mbf{0.72 \pm 0.09}$  \\\hline
			\DPDNetFast   & $99.43$         & $97.54$       & $98.47       \pm 0.13$ &  $2.46$       &  $0.34$       &        $1.14 \pm 0.12$  \\\hline
			\WaterFilling & $93.68$         & $99.02$       & $96.27       \pm 0.21$ &  $0.98$       &  $4.08$       &        $2.91 \pm 0.18$  \\\hline
			\SLICEPCA{}*  & $91.52$         & $92.14$       & $91.83       \pm 0.30$ &  $7.86$       &  $5.20$       &        $6.20 \pm 0.26$  \\\hline
			\SLICESVM{}*  & $90.85$         & $92.22$       & $91.53       \pm 0.30$ &  $7.78$       &  $5.66$       &        $6.46 \pm 0.27$  \\\hline
			\MexicanHat   & $91.81$         & $71.40$       & $80.33       \pm 0.43$ & $28.60$       &  $3.89$       &        $13.27 \pm 0.37$  \\\hline
			\hline
	\end{tabular}}
\end{table*}

Figure~\ref{fig:mivia-performance} shows a much more significant performance drop
of the non CNN-based methods, with a clear superiority of the \DPDNet{}
and \DPDNetFast{} approaches in $Precision$ and $F1_{score}$. Their
worse performance in $Recall$ as compared with \WaterFilling{} was
already explained in previous sections. In what respect to the error
metrics of Figure~\ref{fig:mivia-error}, the CNN-based methods achieve the best
results in the overall error, again with statistically significant
results as compared with the others.

\begin{figure}[h]
	\begin{center}
		\begin{subfigure}[t]{0.49\textwidth}
			\includegraphics[width=\linewidth]{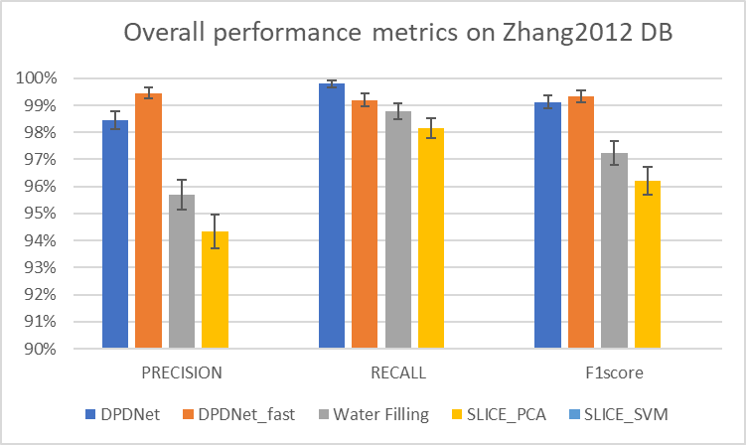}
			\caption{Performance metrics.}
			\label{fig:zhang2012-performance}.
		\end{subfigure}
		\;    \begin{subfigure}[t]{0.49\textwidth}
			\includegraphics[width=\linewidth]{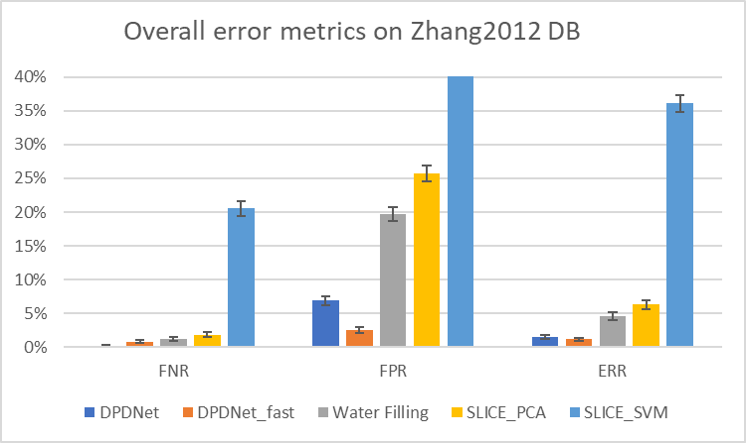}
			\caption{Error metrics.}
			\label{fig:zhang2012-error}.
		\end{subfigure}
	\end{center}
	\caption{Overall performance and error metrics for the \DBZHANG{} dataset.}
	\label{fig:results-zhang}
\end{figure}

Table~\ref{tab:results-zhang} shows the final aggregated results of the evaluated
algorithms on the \texttt{ZHAN2012} dataset, and in
Figures~\ref{fig:zhang2012-performance} and~\ref{fig:zhang2012-error} we show
the graphical comparison among the different strategies, where it is
again clear how our proposals clearly outperform the other algorithms,
with statistically significant improvements. 

\begin{table*}[h]
	\centering
	\caption{Results for the \DBZHANG{} dataset. All values in $\%$.}
	\label{tab:results-zhang}
	\resizebox{\textwidth}{!}{    \begin{tabular}{c|ccc|ccc}
			\hline    \hline
			DB            & $Precision$   & $Recall$      & $F_1score$             &  $FNR$        &   $FPR$        &   $ERR$                  \\\hline\hline
			\DPDNet       & $98.46$       & \cellcolor{222}$\mbf{99.79}$ & $99.12       \pm 0.25$ &  \cellcolor{222}$\mbf{0.21}$ &   $6.87$       &         $1.44 \pm 0.32$        \\\hline
			\DPDNetFast   & \cellcolor{222}$\mbf{99.44}$ & $99.19$       & \cellcolor{222}$\mbf{99.31} \pm 0.22$ &  $0.81$       &   \cellcolor{222}$\mbf{2.47}$ &   \cellcolor{222}$\mbf{1.12 \pm 0.28}$  \\\hline
			\WaterFilling & $95.70$       & $98.79$       & $97.22       \pm 0.44$ &  $1.21$       &  $19.71$       &        $4.61 \pm 0.57$        \\\hline
			\SLICEPCA{}* & $94.35$       & $98.16$       & $96.22       \pm 0.52$ &  $1.84$       &  $25.69$       &        $6.28 \pm 0.66$        \\\hline
			\SLICESVM{}*  & $77.50$       & $79.44$       & $78.46       \pm 1.10$ &  $20.56$       & $110.57$       &       $36.09 \pm 1.29$        \\\hline
			\MexicanHat   & $36.56$       & $23.68$       & $28.74       \pm 1.22$ & $76.32$       & $182.85$       &       $95.87 \pm 0.54$          \\\hline\hline
	\end{tabular}}
\end{table*}

\subsection{Computational Performance Evaluation}
\label{sec:perf-eval}

Regarding the computational demands, Table~\ref{tab:fps-figure} shows the
performance of the evaluated algorithms measured in average frames per
second (we have explicitly removed results for \MexicanHat{} due to its
bad performance and its bad results in computational behavior according
to the results shown in~\cite{luna2017}).

All the experiments shown were run in a standard PC, with an Intel
I7-6700 at 3.4GHz and 32Gb RAM. The GPU used was an NVIDIA GTX 1080
Ti. The evaluation was run on samples from the \DBGOTPD{} dataset.

In the case of the \DPDNet{} and \DPDNetFast{} we also show CPU results.

\begin{table*}[h]
	\centering
	\resizebox{\textwidth}{!}{
		\begin{tabular}{cccccc}
			& \DPDNet{} & \DPDNetFast{} & \WaterFilling{} & \SLICEPCA{} & \SLICESVM{}  \\\hline\hline
			FPS (CPU) & $2.7$     & $9.8$         & $16.6$          & $312.0$      & $406.0$      \\\hline   
			FPS (GPU) & $65.4$    & $116.9$       &     N/A         &  N/A        &  N/A         \\\hline    \hline
	\end{tabular}}
	\caption{Timing Results, measured as average frames per second (FPS).}
	\label{tab:fps-figure}
\end{table*}

As it can be clearly seen, the computational performance of the DNN
based strategies running on a generic CPU is well behind that obtained
by the \WaterFilling{} and \texttt{SLICE} approaches, although the
\DPDNetFast{} could run at a reasonable rate of 10 FPS. The performance
of the \texttt{SLICE} algorithms is much better than the one reported
in~\cite{luna2017} as the CPU we use is significantly better. When the
GPU enters the competition, the results for \DPDNet{} and \DPDNetFast{}
are, as expected, much better than those for \WaterFilling{}, but still
behind \SLICESVM{} and \SLICEPCA{}. However both DNN-based strategies
are well above the requirements demanded for real time operation. Taking
into account that these approaches significantly outperform the other
methods, there is still room for further reduction in the network
complexity.

In what respect to the effect of the scene complexity on the
computational performance of the algorithms, \cite{luna2017} already
showed that the \SLICEPCA{} algorithm was affected by the number of
persons in the scene (this was due to the fact that the ROI estimation
and the feature extraction modules, which demands that are proportional
to the number of people, had a big impact in the processing time). On
the other hand, the \DPDNet{}, \DPDNetFast{}, and \WaterFilling{}
algorithms do not exhibit such a scene complexity
dependence.Table~\ref{tab:fps-numusers} shows the FPS performance for
these algorithms evaluated on sequences with one, two, and more than two
people. It can be seen that there is no direct impact of the number of
persons on the results for \DPDNet{}, \DPDNetFast{}, and
\WaterFilling{}. For the \SLICEPCA{} and \SLICESVM{} approaches, the
relative decrease in computational performance between the FPS for the
single and more than two people cases is of $65.3\%$ and $58.4\%$,
respectively.

\begin{table*}[h]
	\centering
	\resizebox{\textwidth}{!}{
		\begin{tabular}{cccccc}
			& \DPDNet{} (GPU) & \DPDNetFast{} (GPU) & \WaterFilling{} & \SLICEPCA{} & \SLICESVM{} \\\hline\hline
			Single person        & $66.9$          & $111.5$             & $16.6$          & $572.0$     & $682.0$     \\\hline   
			Two people           & $60.9$          & $124.0$             & $16.8$          & $391.8$     & $498.8$     \\\hline    \hline
			More than two people & $65.5$          & $117.5$             & $16.5$          & $198.7$     & $283.4$     \\\hline    \hline
	\end{tabular}}
	\caption{Average timing results for the \WaterFilling{}, \DPDNet{} and
		\DPDNetFast{} algorithms for sequences with varying number of users (FPS).}
	\label{tab:fps-numusers}
\end{table*}

\section{Conclusions}
\label{sec:conclusions}

In this work we propose two specific neural networks (\DPDNet{} and
\DPDNetFast{}) that detect multiple people from depth data captured by
an overhead ToF sensor.  We evaluate and compare our method with other
recent state-of-the-art methods using three different depth image
datasets and following a rigorous experimental procedure. We took into
account that the training and testing subsets were fully independent,
and that the scenes were as realistic as possible, that is, that they
contained people of different heights, with different hairstyles,
wearing different,talking on a mobile phone, and that there were objects
such as different types of chairs.

Our proposal outperforms the other methods in all datasets even
when the number of people is high and they are very close to each
other. Additionally, there was no need to retrain nor fine-tune the
network to the different datasets used.

Comparing the two proposed method, \DPDNetFast{} is $38\%$ faster than
\DPDNet{} at the cost of $0.1\%$ loss decrement in accuracy.  Future
works include adaptation of our approach to work with non-overhead depth
sensors and the use of other network topologies to improve accuracy and
efficiency.

\bibliographystyle{ieee}
\bibliography{DPDnet}

\end{document}